# MetricalARGS: A Taxonomy for Studying Metrical Poetry with LLMs

**Chalamalasetti Kranti**
Department of Linguistics
University of Potsdam, Germany
kranti.chalamalasetti@uni-potsdam.de

**Sowmya Vajjala**
National Research Council, Canada
sowmya.vajjala@nrc-cnrc.gc.ca

## Abstract

Prior NLP work studying poetry has focused primarily on automatic poem generation and summarization. Many languages have well-studied traditions of poetic meter which enforce constraints on a poem in terms of syllable and phoneme patterns. Such advanced literary forms offer opportunities for probing deeper reasoning and language understanding in Large Language Models (LLMs) and their ability to follow strict pre-requisites and rules. In this paper, we introduce MetricalARGS, the first taxonomy of poetry-related NLP tasks designed to evaluate LLMs on metrical poetry across four dimensions: **A**nalysis, **R**etrieval, **G**eneration, and **S**upport. We discuss how these tasks relate to existing NLP tasks, addressing questions around datasets and evaluation metrics. Taking Telugu as our example language, we illustrate how the taxonomy can be used in practice. MetricalARGS highlights the broader possibilities for understanding the capabilities and limitations of today's LLMs through the lens of metrical poetry.

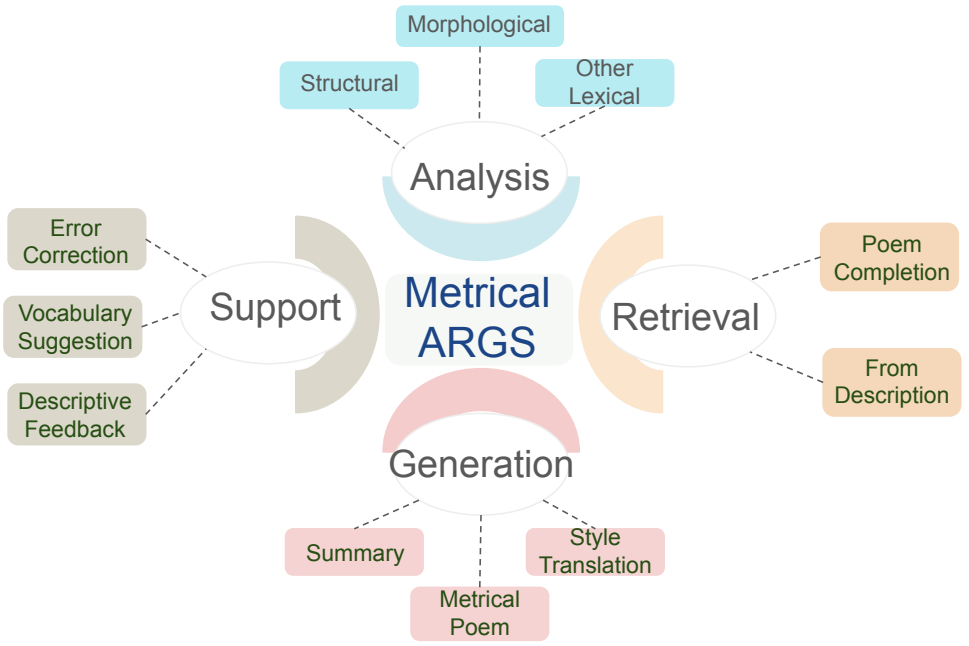

Figure 1: The MetricalARGS taxonomy of tasks for metrical poetry, spanning four dimensions: Analysis, Retrieval, Generation, and Support.

## 1 Introduction

There has been a consistent interest in the generation and analysis of the poetic form in NLP research and other related areas like computational creativity. However, most of this research is primarily focused on English, with some interest in a few other relatively high-resource languages such as Chinese. While automated poetry generation/translation has been the most commonly studied problem, there are also several other related problems when we consider more constrained literary-linguistic systems such as meter, which are governed by their own set of rules and requirements. Each language has specific rules for producing metrical poetry, and has specific dataset needs which has been a bottleneck to extend the studies in computational creativity into other languages.

The advent of LLMs that are capable of in-context learning with very few (or no) examples opens up a possibility of extending the research on computational analysis and generation of poetry to other world languages with rich literary traditions. Many of these poetic traditions involve intricate rules and formal specifications that go beyond free-form generation, which we refer to as metrical poetry throughout this paper. Existing research on related topics in NLP has largely focused on individual tasks in isolation, and there is no unified taxonomy to connect them. In this paper, we take the first steps in addressing this gap by first creating a taxonomy of metrical poetry tasks covering four dimensions - Analysis, Retrieval, Generation, and Support, which we call MetricalARGS (Figure 1). We then demonstrate how to use the taxonomy to study the capabilities of LLMs by considering Telugu, a language with centuries of established metrical poetic tradition (Rao and Shulman, 2020).

From a methodological standpoint, studying metrical poetry with LLMs is important for two main reasons. First, it provides a rigorous testbed for understanding the capabilities of LLMs. Metrical verse requires models to coordinate multiple layers of linguistic competence, including phonology and prosody (to identify and count syllables correctly: a quantitative constraint), morphology and rhythmic structure (to follow metrical and phonological patterns: a structural constraint), and syntax and semantics (to maintain coherence, meaning, and thematic flow: a semantic constraint), all while preserving stylistic and aesthetic consistency. Second, metrical systems are defined by explicit, algorithmic rules that govern syllable patterns, rhyme positions, and line breaks. This makes them inherently computational in nature and opens up opportunities for developing modeling approaches and evaluation methods that can be integrated with mainstream NLP.

Beyond its methodological significance, studying metrical poetry with LLMs also carries important cultural and pedagogical value. Introduction to the poetic meter in Telugu happens in the high school level in the standard educational system where Telugu is the official language, which can also be expected to be a standard practice for other languages with such long poetic traditions. Therefore, exploring the relevance of LLMs for metrical poetry also holds a strong pedagogical potential in supporting student learners as well as adult learners. It could also revitalize interest in a classical literary form of the language, assist with cross-linguistic studies of the poetic form, and support other digital humanities research.

To our knowledge, this is the first paper to propose a unified taxonomy of tasks for metrical poetry exploring how and where LLMs can support them, rather than focusing on one specific task. Further, this is also the first paper that assess LLMs on Telugu metrical poetry.

With this motivation, we make the following contributions in this paper:

1. We describe how LLMs can potentially be useful in the generation and analysis of metrical poetry, by creating a taxonomy of tasks and connecting them to standard NLP tasks and outline dataset and evaluation considerations for each task (Section 3).

2. Taking Telugu as the example language, we demonstrate how LLMs can be used for each of these tasks. Our case study (Section 5) serves as an illustrative probe, and identifies the potential and limitations of using LLMs for metrical poetry related tasks.

## 2 Related Work

Most work related to the poetic form in NLP research has focused on poem generation (Ghazvininejad et al., 2016; Gonçalo Oliveira, 2017; Lau et al., 2018; Van de Cruys, 2020; Ormazabal et al., 2022) including recent research involving LLMs (Belouadi and Eger, 2023; Yu et al., 2024; Qu et al., 2025; Koziev and Fenogenova, 2025). In terms of the studied languages, English (Chakrabarty et al., 2022; Walsh et al., 2024), Chinese (Pan et al., 2023; Ma et al., 2023), and Arabic (ElOraby et al., 2022; Alghallabi et al., 2025) are most commonly studied, with other languages such as Portuguese (Valença and Calegario, 2025) and Russian (Koziev and Fenogenova, 2025) receiving some attention. Tasks such as poetry analysis (Kao and Jurafsky, 2012; Kesarwani et al., 2017; Gopidi and Alam, 2019; Kurzynski et al., 2024) and translation into a given style/language (Genzel et al., 2010; Ghazvininejad et al., 2018; Chakrabarty et al., 2021; Wang et al., 2024) were also explored in the past, and there is a small amount of research on scansion and metrical analysis (Agirrezabal et al., 2017; Valença and Calegario, 2025; Agirrezabal et al., 2016), with some interest in exploring the pedagogical relevance (Zhipeng et al., 2019; Rosa et al., 2025). To our knowledge, individual tasks are considered in isolation so far, disconnected from each other, due to a lack of a common taxonomy.

**NLP, Poetry, and Indian Languages:** To our knowledge, computational analysis of metrical poetry generation/analysis has been studied only for Sanskrit among the Indian languages so far. While some recent research focused on the poetic form, considering aspects such as mood and figure of speech (Sandhan et al., 2025; Jadhav et al., 2025), other research focused on the development of tools such as Chandojnanam (Terdalkar and Bhattacharya, 2023) that scansion a poem i.e., identify the metrical patterns in a poem through rules. Recently, Jagadeeshan et al. (2025) described a dataset and method for poetry generation in Sanskrit using English description as an input, but it primarily focuses on a single metrical form.

Despite consistent academic interest in this topic within the NLP community, there has been no categorization of its specific sub-tasks. This paper addresses that shortcoming and introduces a new language, Telugu, into this line of research.

## 3 Metrical Poetry and NLP Tasks

Meter in poetry can be described as a controlled linguistic system that provides a rhythmic structure to the poems. Meters are typically characterized by rules governing the syllabic and/or sound patterns of the words in a poem, which control the eventual makeup of the poem. We will use syllable based meters as our use case for the rest of this paper. Many languages of the world have established metrical traditions, but the typical process of creating a metrically compliant poem across languages consists of similar steps such as: choosing a metrical form, composing lines that fit the meter, abiding by its rhyme and pattern restrictions, and achieving some form of balance between form and meaning.

In this section, we identify a task taxonomy related to understanding and producing metrical poems (Figure 1) and relate these tasks to the standard tasks studied in NLP research (see Figure 2), as that would enable us to define standard metrics for training and evaluation. We identify four broad categories of tasks: *Analysis, Retrieval, Generation and Support*, and describe them further, looking into potential means of dataset construction and evaluation. Since all of them require deeper understanding and reasoning about the linguistic structure, we do not include reasoning as a distinct task.

### 3.1 Analysis

Analysis tasks concentrate on taking an existing poem and studying its characteristics. We classify analysis into three sub-groups: structural analysis, morphological analysis and other analysis, which are explained below.

**Structural Analysis:** This refers to the poem's adherence to a specific metrical form, and can be split into two broad tasks:

1. **Syllabification and Syllable Classification**: Given a line from a poem, or the full poem, the task is to identify the correct syllable pattern (or other phonetic length patterns, in non-syllabic meters) in the word sequence, and group them together into the appropriate syllable sequences. For example, in Indian languages such as Sanskrit and Telugu, syllable groups are typically 1–4 syllables long, characterized by heavy (*guru*) and light (*laghu*) syllables.
2. **Mapping a syllable pattern to a meter**: Scansion is the process of identifying the metrical structure of a verse, which involves multiple sub-tasks such as counting the syllables, determining rhyme and other pattern based rules, and verifying them with the rules of the available meters to assign the given pattern to a meter.

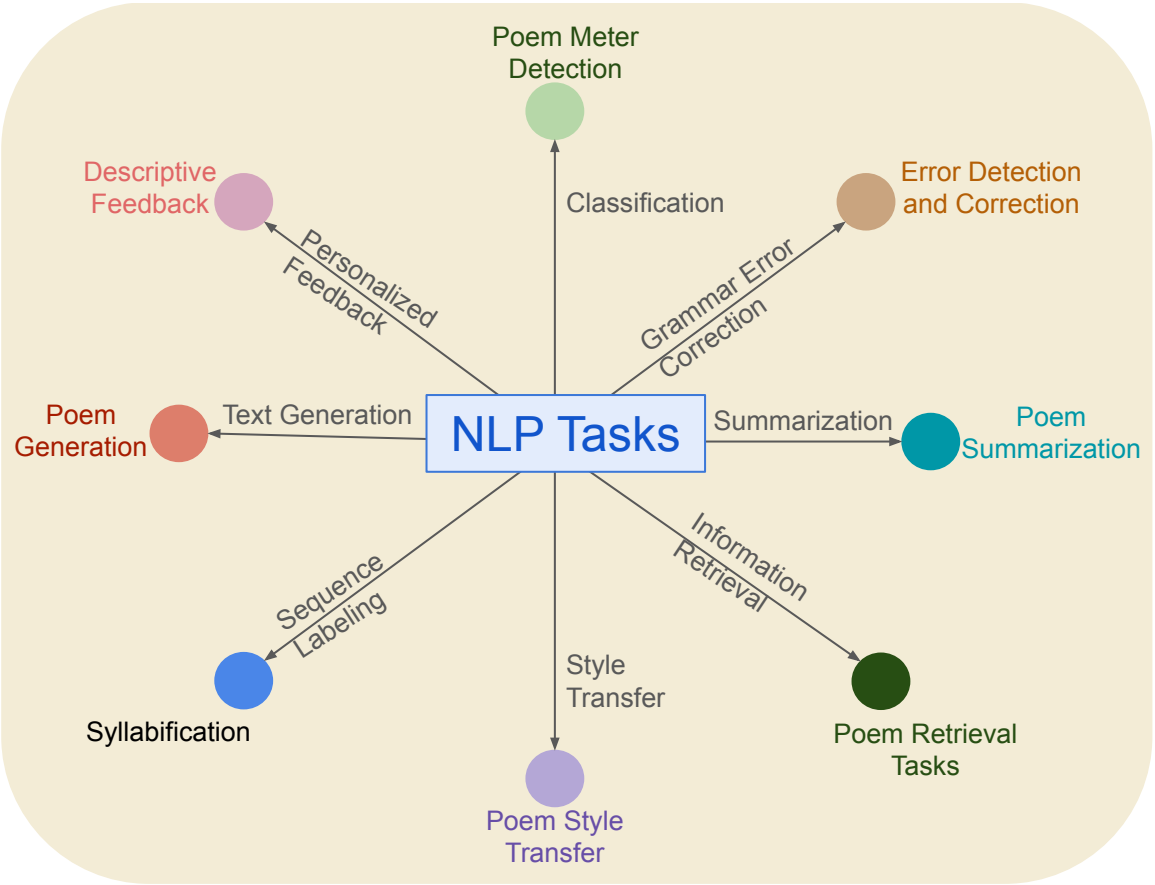

Figure 2: Mapping between METRICALARGS subtasks and established NLP tasks. Each poetry-specific task (outer nodes) aligns with a well-known NLP task (center), illustrating how METRICALARGS builds on existing NLP paradigms while extending them to metrical poetry.

The first task can be compared to a standard sequence labeling task such as part-of-speech tagging or named entity recognition in NLP, and the evaluation measures inherited from sequence labeling research can be adapted to this task. Metrical verse identification is a classification problem, as the number of known metrical patterns is a fixed number. Thus, both the tasks can be measured in terms of accuracy of some form i.e., the number of correctly identified patterns/meters. In terms of data collection, while there are no available datasets of this nature, many classical poems are published with an indicator that identifies their meter. Third party rule-based metrical analysis software already exist for some Indian languages such as Telugu[1], Sanskrit[2], which can be used to build a large scale dataset to support a larger evaluation or fine-tuning of the structural analysis abilities of LLMs.

**Morphological Analysis:** This refers to tasks related to glossing of a given poem, which typically involves breaking up the individual words and adding their meanings in a more colloquial language. In languages with word compounding (many Indian languages, for example), it is a non-trivial process to achieve the appropriate split, and since words can have multiple senses and meanings in context, the task of mapping a word to its right meaning also would involve some form of reasoning. There may also be additional information provided in such glosses, such as person/number/tense information etc. While there are no existing NLP datasets for this task, there are several publicly accessible texts with commentaries on classical metrical poems that contain a gloss of the poem, which can be utilized to build a dataset for this task. There is an already existing body of work on glossing in NLP (Ginn et al., 2023), and evaluation measures from that research can be easily adapted to this task.

[1] https://chandamu.github.io/
[2] https://sanskrit.iitk.ac.in/jnanasangraha/chanda/

**Other Analysis:** In digital humanities, as well as in NLP, it is not uncommon to see research studying sentiment, lexical/syntactic/stylistic patterns, authorship attribution and so on. both for prose and poetry. Hence, it is natural to study these tasks to understand the capabilities of LLMs in this area. However, while evaluation may be straightforward as such analysis may easily fit into a standard text classification task framework, compilation of relevant datasets for each of these tasks would require some human expertise. The role of LLMs in supporting humans in building high quality datasets for this kind of problems can also be explored in this context.

### 3.2 Retrieval

We refer to tasks related to identifying the right (existing) poems based on user queries as retrieval, which are listed below:

1. Retrieving the poem given its starting words or the first verse or words that appear in the middle or at the end
2. Retrieving the poem from its description
3. Retrieving the poem(s) that matches in meaning, meter etc.

All of these tasks are similar to search and information retrieval tasks. Considering the large body of classical poetry based literature already available online, collecting datasets at least for the first two of these retrieval tasks should be straightforward. Since there are existing compilations of poem-summary pairs for some languages such as Telugu,[3] they can be utilized as a starting point for creating a larger scale dataset, potentially with synthetically generated paraphrased versions of summaries, which can be useful for evaluation and fine-tuning purposes. In terms of evaluation, the standard retrieval based evaluation measures such as precision/recall/F-score can be used.

### 3.3 Generation

Generation tasks involve some form of textual generation based on a given description. We identify three generation tasks, described below:

1. **Poem summarization:** translating a metrical poem, typically written in the classical literary form of the language into plain text. This can be viewed as a task similar to text summarization which is well-studied in NLP.
2. **Poem generation:** generating a novel poem given a textual description and a specified meter. Poem generation under such linguistic constraints is also a explored in the past in NLP, especially for English and Chinese languages.
3. **Poem style transfer:** This is a challenging variation to generating a novel poem, where the input is a poem in one meter, and the output is the same content adapted to another meter. This task aligns with other existing research on style transfer in NLP, but in the context of poetry.

For summarization, it is easier to create larger scale datasets by tapping into available resources, but for the other generative use cases, which explore novel and creative content generation, one option is to look at synthetic data generated from LLMs followed by human evaluation. The effectiveness of generation can be evaluated through standard text generation metrics and metrical adherence can be checked through rule based checkers. But, human ratings are a must for other factors such fluency, coherence, adherence to the theme and style, and aspects of creativity. Data creation and evaluation are perhaps the most challenging for these generative tasks compared to all the other tasks discussed in this paper.

### 3.4 Support

Support tasks explore the role of LLMs in offering support to poets and students learning to write

[3] https://huggingface.co/datasets/SuryaKrishna02/aya-telugu-poems

metrical poetry. We identify three main tasks as a starting point.

1. **Error detection and correction:** This refers to the process of identifying metrical, lexical or grammatical errors in the user written poem and offering corrections. This is most similar to the grammatical error detection/correction tasks, and word-level translation quality estimation tasks, that have a long history in NLP research.

2. **Vocabulary suggestion:** This task, as the name indicates, offers vocabulary suggestions, but aligning with the metrical constraints of the context. There is perhaps no equivalent existing NLP task as these suggestions need both semantic and metrical compliance.

3. **Descriptive Feedback:** This refers to giving explanation to the user on the text they created and offering suggestions for rewriting. The more recently introduced Grammatical Error Explanation task (Song et al., 2024) is potentially the closest existing NLP task.

All the three support tasks, while being specific to metrical poetry generation, also share some commonalities with other relevant research on educational applications of NLP, and data creation and evaluation approaches for the related topics can be adapted for these use cases as well. However, they are inherently more challenging tasks than the other ARGS tasks.

From this discussion, it is clear that most of the METRICALARGS tasks can benefit from existing research in related tasks, while introducing challenging new variations. Overall, the METRICALARGS taxonomy demonstrates that there are a wide range of complex tasks related to metrical poetry generation, where LLMs may be relevant and can be further studied. While the taxonomy is created with Telugu meter as its basis (owing to the authors' familiarity with it), these tasks are not specific to Telugu and can be studied for other languages with similar poetic traditions as well. We expect this task taxonomy to be adapted and improved as needed for developing similar task classifications for other languages and language families.

Figure 2 summarizes the different METRICALARGS tasks and their relation to other standard NLP tasks. Table 1 shows how some of the existing research maps into this taxonomy. Most of the past work appears to have focused on Analysis and Generation tasks, often addressing only a subset of sub-tasks within each category, while Retrieval and Support remain largely unexplored.

| Paper | Lang | Task | METRICALARGS |
|---|---|---|---|
| Kao and Jurafsky (2012) | en | PA | Analysis |
| Walsh et al. (2024)* | en | PF | Analysis |
| Valença and Calegario (2025)* | pt | PS | Analysis |
| Pan et al. (2023) | zh | PA | Analysis |
| Kurzynski et al. (2024) | zh | PP | Analysis |
| Jagadeeshan et al. (2025) | sa | PG | Generation |
| Ghazvininejad et al. (2016) | en | PG | Generation |
| Belouadi and Eger (2023)* | en | PG | Generation |
| ElOraby et al. (2022) | ar | PG | Generation |
| Koziev and Fenogenova (2025)* | ru | PG | Generation |
| Genzel et al. (2010) | en | ST | Generation |
| Wang et al. (2024)* | en | ST | Generation |

Table 1: Mapping existing metrical poetry works with the proposed METRICALARGS tasks. Lang: Languages supported; PA: Poetry Analysis; PF: Poetic Form; PS: Poetic Scansion; PP: Poem Parllelism; PG: Poem Generation; ST: Style Transfer; * - Indicates the works used LLMs.

## 4 Applying METRICALARGS for Telugu

We demonstrate the use of METRICALARGS taking Telugu as the test language in this section. Telugu is recognized as one of the classical languages of India (Press Information Bureau, 2024) and has a centuries old literary tradition. Earliest known description of Telugu poetic meters and rules of prosody are from a 6th or 7th century text (Ramakrishna et al., 1983, pp.164–165). Telugu poetic meter, while sharing a lot of patterns with Sanskrit poetic tradition (*chandas*) has a lot of other native metrical patterns as well. Considering the agglutinative nature of the language, tasks such as breaking up of the syllable sequences into individual words for glossing and summarizing the meaning too offer a range of language processing and reasoning related challenges, along with other tasks around metrical poetry analysis and generation.

To illustrate the use of METRICALARGS taxonomy for Telugu metrical poetry, we curated a pilot dataset of approximately 20 samples for each task (170 samples in total). The intention of using such a small dataset is not to establish a benchmark, but to showcase how current models handle ARGS tasks in the Telugu language and illustrate how to build benchmark datasets using this taxonomy across languages in future. Although modest in size, the dataset covers representative examples of each task. These samples were collected from the official Grade 7–10 Telugu textbooks published by the Andhra Pradesh state government in India. The dataset was prepared and annotated by two native Telugu speakers.

| Category | SubCategory | # Q | Accuracy | |
|---|---|---|---|---|
| | | | GPT | Gemini |
| Analysis | SC | 20 | 0.60 | 0.20 |
| | MA | 20 | 0.20 | 0.65 |
| | MD | 20 | 0.40 | 0.50 |
| Retrieval | FV | 6 | 0.00 | 0.00 |
| | MRV | 6 | 0.00 | 0.00 |
| | LV | 6 | 0.00 | 0.00 |
| Generation | PS | 20 | 0.70 | **0.85** |

Table 2: Accuracy scores across different MetricalARGS tasks. SC: Syllabification and Syllable Classification, MA: Morphological Analysis, MD: Meter Detection, FV: Retrieval from First Verse, MRV: Retrieval from Middle/Random Verse, LV: Retrieval from Last Verse, PS: Poem Summarization. #Q indicates the number of questions per task.

| Category | SubCategory | # Q | Accuracy | |
|---|---|---|---|---|
| | | | GPT | Gemini |
| Retrieval | MM | 2 | 0.50 | 0.00 |
| Generation | PFS | 7 | 0.71 | 0.29 |
| | RPFW | 8 | **1.00** | 0.75 |
| | PFP | 5 | 0.40 | 0.20 |
| | ST | 20 | 0.00 | 0.00 |
| Support | EDC | 10 | 0.10 | 0.00 |
| | VS | 19 | 0.47 | 0.26 |

Table 3: Accuracy scores for tasks evaluated using an LLM-as-a-judge. MM: Retrieval using meaning, PFS: Poem from Summary, RPFW: Riddle Poem from Word, PFP: Poem from a Problem (Samasya in Avadhanam), ST: Style Transfer, EDC: Error Detection and Correction, VS: Vocabulary Suggestion. #Q indicates the number of questions per task.

**Evaluation** We considered two proprietary LLMs for output generation: GPT-5 and Gemini-2.5-Pro. For the analysis tasks (syllabification, syllable classification, morphological segmentation, and meter validation) as well as one generation sub-task (summarization), gold references were available. For the remaining categories, such as retrieval, poem generation, and style transfer, multiple valid outputs are possible. In both the cases, we adopted an LLM-as-a-judge approach for evaluation and used Gemini-2.5-Pro as the judge model, supplying it with a gold output for comparison where it is available. The percentage of responses the judge model scores as correct is considered as the measure of performance (with a scale of 0–1, the more the better). All experiments were conducted using the Inspect LLM evaluation framework[4] in a zero-shot setting, using Telugu prompts (in Appendix A.1), with temperature set to zero. We also conducted a human evaluation in which authors, who are native Telugu speakers reviewed a sample of outputs to verify both the correctness of model predictions and the validity of the LLM-based evaluations [5].

## 5 Results

Table 2 presents the results for the tasks where gold references are available. Among these, for poem summarization task, both GPT-5 and Gemini-2.5-Pro performed well, reporting overall scores of 0.70 and 0.85 respectively as per the judge model. While the GPT-5 model did better with syllabification, Gemini model did better with morphological analysis. Meter detection was more challenging for both models, though. The models often misclassified syllable length, leading to downstream errors in meter identification. In morphological analysis, both the models generally captured meanings with multiple words instead of single-word glosses, resulting in mismatches with gold annotations, indicating the need for better evaluation measures for that task. Surprisingly, retrieval proved to be a challenging task. Neither model was able to successfully retrieve the complete poem given only the first, last, or a random line as input, resulting in an accuracy of 0 for this task. GPT-5 avoided retrieval altogether by producing follow-up questions (see Figure 3), while Gemini-2.5-Pro generated paraphrased versions of the poems rather than exact matches. This could be due to potential coverage issues in the training data, or the models trying to avoid verbatim reproduction of text.

Table 3 reports the results for the tasks where the LLM judge evaluated the generated outputs without a gold answer. GPT-5 appears to be better than Gemini for these tasks in general, while both models achieve zero or near-zero scores for style transfer and error detection and correction. Overall, these results provide preliminary insight into how models approach different categories of metrical tasks, highlighting areas of relative strength (semantic-level tasks) and weakness (retrieval and structural-based tasks). Note that this evaluation covered all the tasks listed in the taxonomy, with two exceptions: "Other Analysis" subset of tasks require expert data curation, and "Descriptive Feedback" is assessed through its components (er-ror correction and vocabulary suggestion). Hence, they are excluded from this case study.

[4] https://inspect.aisi.org.uk/

[5] The dataset and the human evaluation sample will both be provided as supplementary material upon acceptance.

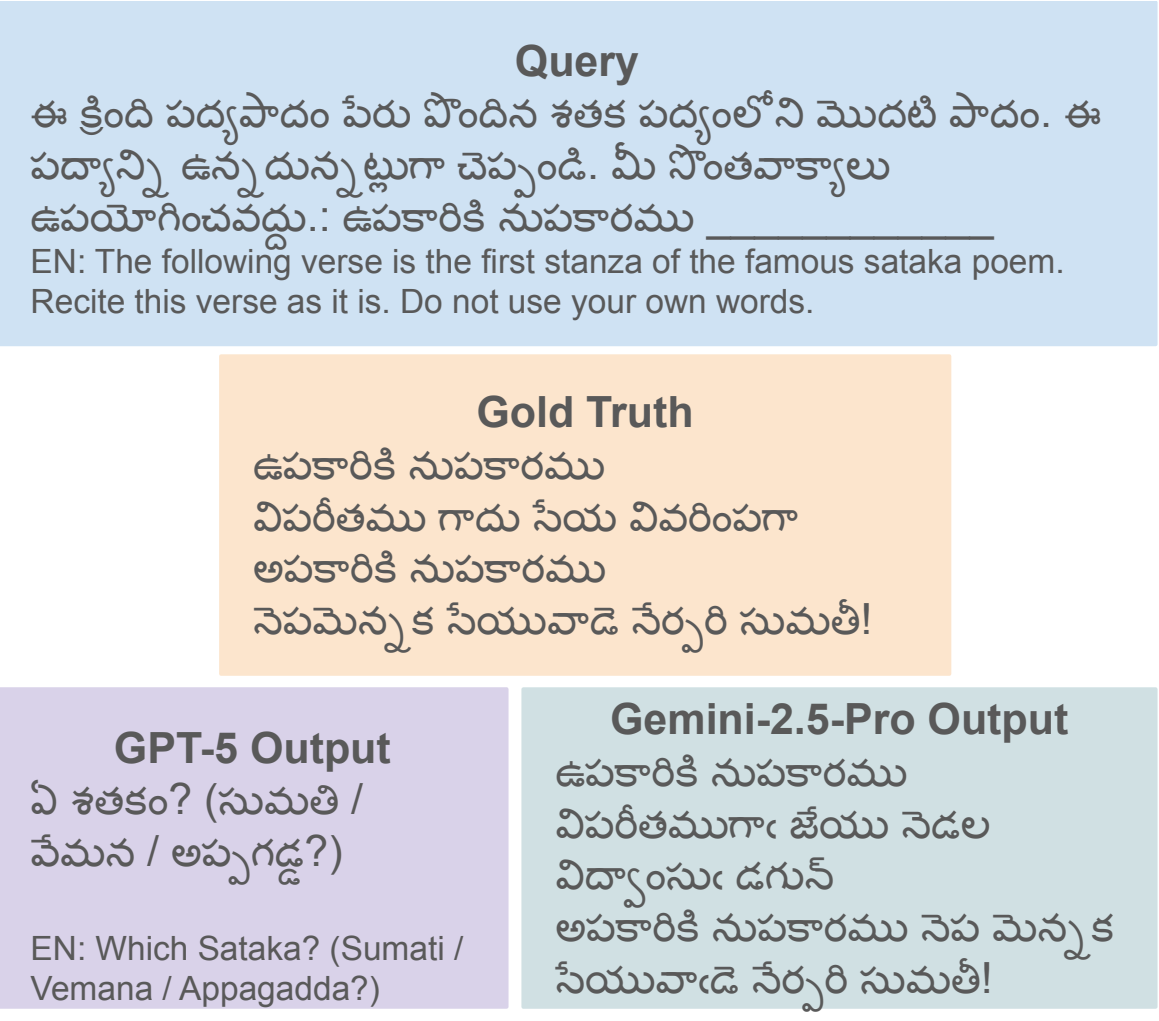

Figure 3: Example from the Retrieval QA task. GPT-5 responded with a clarification question instead of retrieval, while Gemini-2.5-Pro produced text with partial overlap, but did not match the gold reference.

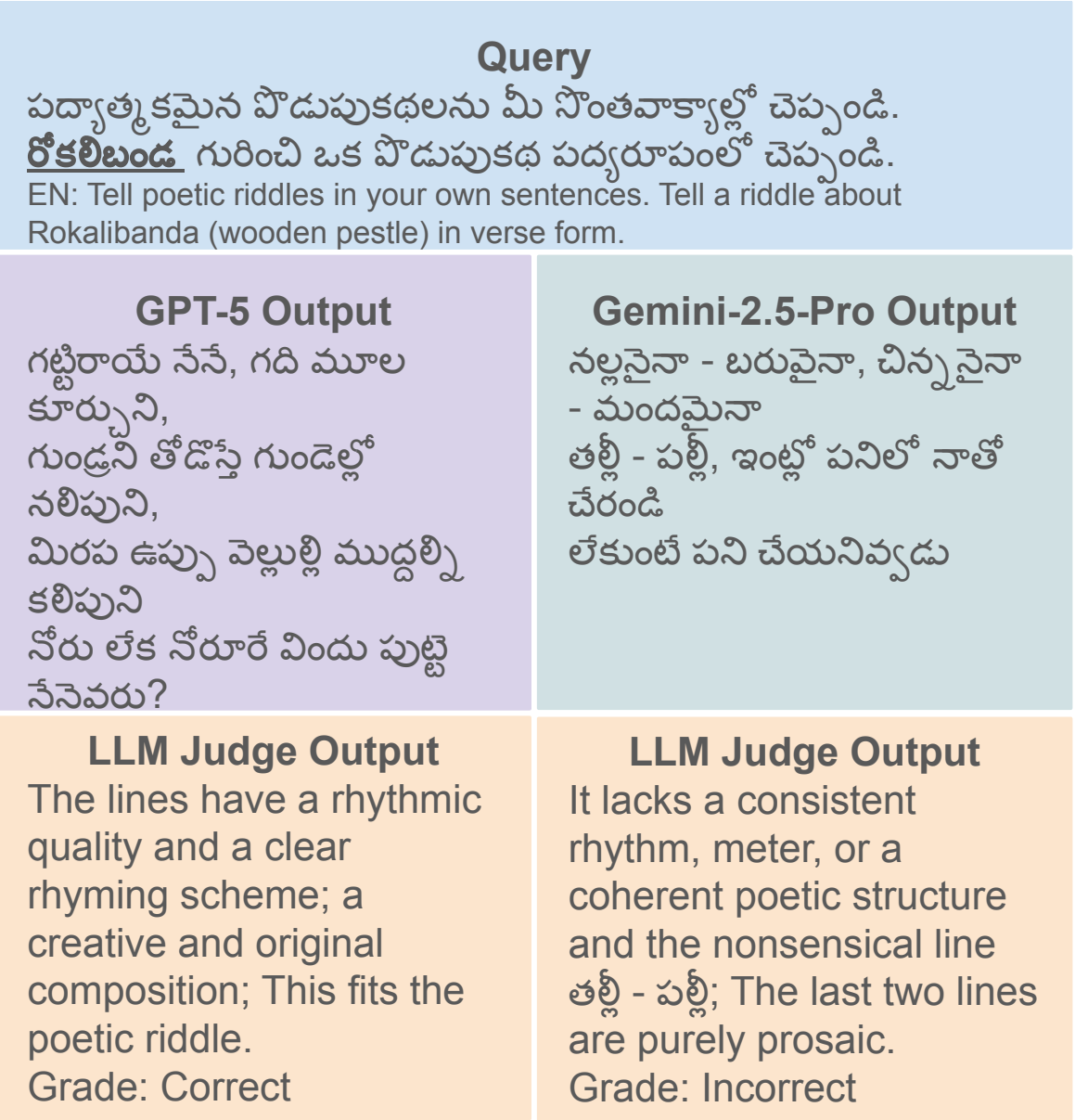

Figure 4: Example from the Riddle-style Poem Generation task. The query asks the model to compose a riddle in verse form about Rokalibanda (a wooden pestle).

### 5.1 Qualitative Analysis

To better understand model behavior, we examine two representative tasks focusing on retrieval and generation respectively. Figure 3 illustrates a retrieval question task where the query provides a first verse and the model is asked to retrieve the complete poem, while the gold standard answer has the expected poem. GPT-5 responded with a follow-up question rather than attempting a retrieval; this behavior was consistent across all samples in this task. Gemini-2.5-Pro produced outputs with some lexical overlap with the gold poem but did not match the target verse, and therefore did not satisfy the exact-retrieval criterion. Manual analysis revealed a consistent pattern in how the Gemini-2.5-Pro model used the input verse. When given the first or last verse of the poem, the model retained the input at the appropriate position and attempted to generate surrounding verses. However, the same behavior extended even to the random (middle) verse case, where the input fragment was incorrectly treated as the opening line, and the model generated a continuation around it. This suggests that rather than recalling the original poem, the model interprets the task as one of conditional generation. This task probes the model's ability to match patterns and recall memorized sequences. Given that LLMs generate text through next-token prediction, one might expect them to reproduce a familiar verse when prompted with part of it. However, in our experiments, neither model succeeded, indicating unreliable recall of longer structured texts. Possible reasons include safety guardrails that prevent verbatim reproduction and limited Telugu training coverage, warranting further tests in other languages.

Figure 4 presents the model outputs and the LLM-judge's reasoning criteria for a poem generation question. In this example, the model is asked to generate a poem in the form of a riddle for a given word. To succeed, the model must first understand the meaning of the word and identify its characteristic features, and then compose a poem that conveys those features indirectly, without explicitly naming the word. This combination of semantic understanding, abstraction, and structured poetic composition makes the task challenging (even for humans). While the LLM-as-a-judge marked the GPT-5 output as correct, human inspection revealed a mismatch: the verse was rhythmically consistent but contained lexical errors and ungrammatical phrasing in Telugu. This highlights both the complexity of the task and a discrepancy between LLM-based and human judgments of fine-grained linguistic accuracy. Taken together, these examples illustrate that while the LLMs can produce outputs that appear fluent or partially aligned with expectations, they often fail to meet stricter criteria such as exact retrieval or grammatical accuracy, underscoring the need for human verification

| Category | Sub-Category | JS | A1 | A2 |
|---|---|---|---|---|
| **GPT-5** | | | | |
| Retrieval | MM | 0.50 | 0.00 | 0.0 |
| Generation | PFS | 0.71 | 0.29 | 0.43 |
| | RPFW | **1.00** | **0.00** | **0.00** |
| | PFP | 0.40 | 0.20 | 0.20 |
| | ST | 0.00 | 0.00 | 0.00 |
| Support | EDC | 0.10 | 0.00 | 0.00 |
| | VS | 0.47 | 0.26 | 0.26 |
| **Gemini-2.5-Pro** | | | | |
| Retrieval | MM | 0.00 | 0.00 | 0.0 |
| Generation | PFS | 0.29 | 0.14 | 0.14 |
| | RPFW | **0.75** | **0.63** | **0.50** |
| | PFP | 0.20 | 0.20 | 0.20 |
| | ST | 0.00 | 0.00 | 0.00 |
| Support | EDC | 0.00 | 0.00 | 0.00 |
| | VS | 0.26 | 0.11 | 0.16 |

Table 4: A Comparison of Human and LLM judge evaluations. MM: Retrieval using meaning, PFS: Poem from Summary, RPFW: Riddle Poem from Word, PFP: Poem from a Problem (Samasya in Avadhanam), ST: Style Transfer, EDC: Error Detection and Correction, VS: Vocabulary Suggestion.

in addition to LLM-based evaluation.

### 5.2 Human Evaluation of Model Outputs and Judge Scores

As there are discrepancies between model outputs and the LLM-judge assessments, we conducted a human evaluation for a reliable reference. The authors, both native Telugu speakers, independently evaluated the model outputs and marked whether each output was correct. These computed scores are then compared with the scores assigned by the LLM-judge in order to assess both the correctness of the model outputs and the reliability of the LLM-as-a-judge framework. Table 4 shows a summary of this comparison.

Overall, the LLM-as-a-judge (JS) reports higher scores than human annotators (A1 and A2) across most sub-categories (see Table 4), especially for GPT-5. In the Retrieval–MM task, for example, GPT-5 reproduced the input poem from Vemana Satakam with only the closing line altered, rather than retrieving a semantically similar poem (see Figures 8 in the Appendix). The judge incorrectly labeled this as correct, while both the human evaluators marked it as incorrect. In the Generation–Poem from Summarization and Poem (riddle) tasks, GPT-5 outputs often preserved meaning but lacked poetic structure, coherence, or riddle form (see Figure 9 and 10 in the Appendix). The judge, focusing mainly on semantic similarity, labeled it as correct, unlike the human annotators.

In style transfer tasks, outputs achieved partial alignment (upto 70% similarity for some questions) with the target style (see Figure 11 in the Appendix), and occasional deviations, such as GPT-5 producing Sanskrit or Gemini-2.5-Pro generating multiple responses, showing inconsistencies in task adherence. The judge also penalized a correct output for not reproducing a well-known poem, despite the prompt not specifying such a requirement. Overall, these findings indicate that the LLM-as-a-judge relies heavily on meaning-based evaluation, underestimating stylistic, structural, and contextual alignment, highlighting the need for richer, context-aware evaluation frameworks. More experiments with judge prompts, and potentially fine-tuned judge models may be needed for better automated evaluation of such tasks.

## 6 Road Ahead

In this paper, we created a taxonomy of tasks METRICALARGS for metrical poetry analysis with LLMs, and demonstrated how this can be applied to one language, Telugu, by exploring the capabilities of today's LLMs across these tasks. The case study demonstrates both the potential and challenges with the current LLMs. We believe that positioning metrical poetry as a testbed opens up new ways to assess and enhance LLM understanding of form-constrained language, while also supporting application areas such as learning tool development and digital humanities research.

Creation of high quality evaluation datasets and exploring mechanisms to create larger scale fine-tuning datasets across different tasks is an obvious next step to pursue in this direction, to operationalize the ideas described in this paper in practice. Extending to other world languages with similar poetic traditions, and adapting the taxonomy accordingly should be considered too. We hope that this paper serves as a starting point for further research on exploring the relation between LLMs and metrical poetry traditions of other Indian languages, and more broadly, investigating the role of LLMs in understanding other structured linguistic systems like meter across the world languages.

## Limitations

The paper focused primarily on building a conceptual taxonomy and since we used Telugu metrical poetry tradition as the basis for building this taxonomy, the coverage of tasks may not be comprehensive. For example, it is possible to imagine a task like metrical poem translation between languages (*Translate a given poem in Telugu meter T-A to Chinese meter C-X*), which this taxonomy does not cover at the moment. This can be perceived as a limitation, but we intend for this paper to be a starting point to raise further discussions on the topic, and hence, we could expect to see improvements to the taxonomy in near future.

## Ethics and Broader Impact Statement

We don't foresee any particular ethical concerns with this paper, as we did not work with human subjects and took the textual examples used from public domain data.

## Acknowledgments

We thank Isar Nejadgholi, Krishnapriya Vishnubhotla and Gabriel Bernier-Colborne for their feedback. Dileep Miriyala's Chandam (https://chandamu.github.io/) inspired us to study metrical poetry with LLMs, and we thank him for creating the tool.

# A Appendix

## A.1 Prompts

The Telugu Prompt (see Figure 5, 6, and Table 5 ) follows standard zero-shot prompting structure comprising task description. We follow the default template of the Inspect framework to Judge Prompt to evaluate the model generated responses (see Figure 7).

## A.2 Qualitative Analysis

In the Retrieval–MM task, lower human scores for GPT-5 stem from its failure to retrieve a poem with similar meaning from Vemana Satakam. The model reproduced the input poem verbatim with only the makutam (closing verse) changed (see Figure 8), although such a poem does not exist in Vemana Satakam. The LLM-as-a-judge, however, marked this as correct, increasing the score relative to human assessments.

In the Generation–Poem from Summarization task, differences between judge and human scores arise from contrasting emphasis on meaning versus form. GPT-5 outputs generally preserve semantic content but often lack poetic coherence or structure, resembling prose or using contextually misplaced words (see Figure 9). The judge, focusing on semantic similarity, overlooks these stylistic deficiencies and assigns higher scores.

A similar pattern appears in the Generation–Poem (riddle) task, where GPT-5 responses frequently lack logical or poetic structure (see Figure 10) but are still rated correct by the judge. This behavior likely results from the judge evaluating outputs primarily through translation and meaning comparison rather than structural or creative alignment. Consequently, outputs that align in surface meaning but fail in form are treated as correct, while human evaluators apply stricter criteria.

In style transfer tasks, model outputs often achieve partial alignment with the target style, sometimes exceeding 70% similarity (see Figure 11).

Overall, these findings suggest that the LLM-as-a-judge primarily emphasizes semantic similarity while neglecting stylistic, structural, and contextual aspects that human evaluators recognize. This tendency results in higher scores for outputs that align in meaning but lack linguistic or creative quality, highlighting the need for evaluation frameworks that account for contextual and stylistic depth.

TEMPLATE A.2.1
పద్యలక్షణాలను తెలియచేసే శాస్త్రమే ఛందస్సు. ఛందస్సులోని గణాలకు గురు లఘువులు ఆధారం.
లఘువు:
– ఒక మాత్ర (కనురెప్పపాటు కాలంలో లేదా ఒక చిటికె వేసే) కాలంలో ఉచ్చరించే అక్షరాలు
– దీన్ని | తో సూచిస్తారు
– హ్రస్వాచ్చులు (అ, ఇ, ఉ, ఋ, ఎ, ఒ)
– హ్రస్వాచ్చులతో కూడిన హల్లులు (క, చి, గు, బృ, వె, రొ)
– దీర్ఘాలు కాని ద్విత్వాక్షరాలు (క్క, గ్గి, చ్చు, ట్ట, త్తె, బ్బొ)
– దీర్ఘాలు కాని సంయుక్తాక్షరాలు (క్ర, స్యి, ర్థు, శ్మ, ద్వె, త్యొ)
– తేల్చి పలుకబడే రేఫతో కూడిన అక్షరాల ముందు అక్షరాలు (అద్రుచు, కద్రువ, విద్రువ )
– ఉదాహరణ: అ, ఇ, ఉ, ఋ, ఎ, ఒ, క, చి, టె, తు, ప్ృ, జొ, ఘ, ఝ, ఢ, ధ, భ, స, హ
గురువు:
– రెండు మాత్రల కాలంలో ఉచ్చరించే అక్షరాలు
– లఘువు సమయం కంటే ఉచ్చారణకు ఎక్కువ సమయం అవసరమయ్యే అక్షరాలు
– దీన్ని U తో సూచిస్తారు
– దీర్ఘాలు (ఆ, ఈ, ఊ, ౠ, ఏ, ఐ, ఓ, ఔ)
– దీర్ఘాచ్చులు కలిగిన హల్లులు:
U U U U U U U U
కాదు, గీతం, రూకలు, పితౄణం, జేబు, ఖైదు, డోలు, కౌలు)
– ద్విత్వాక్షరానికి ముందున్న అక్షరాలు:
U U U U
అమ్మ, చిల్లర, మబ్బు, కత్తెర
– సమ్యుక్తాక్షరానికి ముందున్న అక్షరాలు:
U U U
పద్యము, భక్తి, కల్పన
– సంశ్లేషాక్షరానికి ముందున్న అక్షరాలు:
U U U
ఉష్ట్రం, వస్త్రము, మర్త్యము
– పూర్ణ బిందువు కలిగిన అక్షరాలు:
U U U
ఛందస్సు, లక్షణం, చెప్పడం
– పొల్లు హల్లుతో కూడిన అక్షరాలు:
U U U
రాజేష్, కొరకున్, ఋత్విక్
– విసర్గ కలిగిన అక్షరాలు:
U U U
దుఃఖం, అంతఃపురం, నమః
– ఉదాహరణ: ఆ, ఈ, ఊ, ఏ, ఐ, ఓ, ఔ, అం, కా, చీ, టూ, తౄ, పే, శై, జో, గౌ, జం, డం, దా
పైన ఇచ్చిన వివరణను ఉపయోగించి, ఈ క్రింది పదానికి లఘువులు, గురువులు గుర్తించండి
$INPUT_QUESTION$

Figure 5: Telugu Prompt template for the syllabification.

TEMPLATE A.2.2
చంపకమాల
– నాలుగు పాదాల పద్యం; ప్రతీ పాదంలోనూ 21 అక్షరాలు; ప్రాస నియమం ఉంది; 11వ అక్షరం యతి స్థానం; న – జ – భ – జ – జ – జ – ర గణాలు వరుసగా వస్తాయి.
ఉత్పలమాల
– నాలుగు పాదాల పద్యం; ప్రతీ పాదంలోనూ 20 అక్షరాలు; ప్రాస నియమం ఉంది; 10వ అక్షరం యతి స్థానం; భ – ర – న – భ – భ – ర – వ గణాలు వరుసగా వస్తాయి.
మత్తేభము
– నాలుగు పాదాల పద్యం; ప్రతీ పాదంలోనూ 20 అక్షరాలు; ప్రాస నియమం ఉంది; 14వ అక్షరం యతి స్థానం; స – భ – ర – న – మ – య – వ గణాలు వరుసగా వస్తాయి.
శార్దూలము
– నాలుగు పాదాల పద్యం; ప్రతీ పాదంలోనూ 19 అక్షరాలు; ప్రాస నియమం ఉంది; 13వ అక్షరం యతి స్థానం; మ – స – జ – స – త – త – గ గణాలు వరుసగా వస్తాయి.
కంద
– నాలుగు పాదాల పద్యం; ఒకటవ మూడవ పాదాల్లోని గణాల సంఖ్య, రెండవ నాల్గవ పాదాలలోని గణాల సంఖ్య సమానంగా ఉంటాయి; ప్రాస నియమం ఉంది; రెండు నాల్గవ పాదాల్లో చివరి అక్షరంగా గురువు ఉంటుంది; ఆరవ గణం జ గణం గాని, నలం గణం గాని ఉంటుంది; బేసి గణాలలో జ గణం రాకూడదు; రెండు, నాలుగు పాదాల్లో 1−4 గణాల మొదటి అక్షరాలకు యతిమైత్రి చెల్లుతుంది; భ – జ – స – న – ల – గగ గణాలు వరుసగా వస్తాయి; మొదటి పాదం గురువుతో ప్రారంభమైతే తక్కిన పాదాలు కూడా గురువుతోనూ, లఘువుతో ప్రారంభమైతే తక్కిన పాదాలు కూడా లఘువుతోనూ ప్రారంభమవుతాయి.
ఆటవెలది
– నాలుగు పాదాల పద్యం; 1వ, 3వ పాదాలు ఒక విధంగాను, 2వ, 4వ పాదాలు ఒక విధంగాను ఉంటాయి; 1, 3 పాదాల్లో వరుసగా మూడు సూర్యగణాలు (న గణము, హ గణము), రెండు ఇంద్రగణాలు (నల – నగ – సల – భ – ర – త) ఉంటాయి; 2, 4 పాదాల్లో ఐదు సూర్య గణాలు ఉంటాయి; ప్రతి పాదంలో నాలుగవ గణంలోని మొదటి అక్షరం యతి స్థానం; యతి లేని చోట ప్రాస యతి చెల్లుతుంది; ప్రాస నియమం లేదు.
తేటగీతి
– నాలుగు పాదాల పద్యం; ప్రతీ పాదంలో వరుసగా ఒక సూర్య గణం, రెండు ఇంద్రగణాలు(నల – నగ – సల – భ – ర – త), రెండు సూర్యగణాలు ఉంటాయి; ప్రతి పాదంలో నాలుగవ గణంలోని మొదటి అక్షరం యతి స్థానం; యతి లేని చోట ప్రాస యతి చెల్లుతుంది; ప్రాస నియమం లేదు.
సీస పద్యం
– ప్రతిపాదం రెండు భాగాలుగా ఉంటుంది; ప్రతి భాగంలో నాల్గేసి గణాల చొప్పున ఒక్కొక్క పాదంలో ఎనిమిది గణాలుంటాయి; పాదం మొదటి భాగంలో 4 ఇంద్రగణాలు, రెండవభాగంలో వరుసగా రెండు ఇంద్రగణాలు, రెండు సూర్యగణాలు ఉంటాయి. ప్రతి పాదంలో 3వ గణంలో మొదటి అక్షరం యతి స్థానం; యతి లేని చోట ప్రాస యతి చెల్లుతుంది; ప్రాస నియమం లేదు.
పైన ఇచ్చిన వివరణను ఉపయోగించి, ఈ క్రింది పద్యం యే ఛందస్సులో ఉందో చెప్పండి.
$INPUT_QUESTION$

Figure 6: Telugu Prompt template for the meter detection questions.

TEMPLATE A.2.3
You are assessing a submitted answer on a given task based on a criterion. Here is the data:
[BEGIN DATA]
***
[Task]:
$INPUT_QUESTION$
***
[Submission]:
$MODEL_OUTPUT$
***
[Criterion]: NA
***
[END DATA]
Does the submission meet the criterion?
After assessing the submitted answer, reply with 'GRADE: $LETTER' (without quotes) where LETTER is one of CI.
Please choose ONE option for the grade: either C̈for correct answers, or Ïfor incorrect answers.
For example, after reviewing a correct answer you might write 'GRADE: C' or after reviewing an incorrect answer you might write 'GRADE: I'.
First, write out in a step by step manner your reasoning about the criterion to be sure that your conclusion is correct. Avoid simply stating the correct answers at the outset. Then, end with your answer formatted as 'GRADE: $LETTER' (without quotes) where LETTER is one of CI.

Figure 7: Default English Prompt template for using LLM as a judge in the evaluation in Inspect Framework.

Q: ఈ క్రింది పద్యభావానికి సరిపోయే వేమన శతకంలోని పద్యాన్ని చెప్పండి. మీ సొంతవాక్యాలు ఉపయోగించవద్దు.
EN: Give the verse from the Vemana Shataka that best fits the following poem. Do not use your own sentences.
నిండు విస్తరాకు నిశ్చలంబుగనుండు
ఏమిలేని విస్తరెగిరిపడును
వినయియగు వివేకి, విఱ్ఱవీగు జడుండు
మాచిరాజుమాట మణులమూట

GPT-5 Output:
నిండు విస్తరాకు నిశ్చలంబుగనుండు
ఏమిలేని విస్తరెగిరిపడును
వినయియగు వివేకి, విఱ్ఱవీగు జడుండు
విష్వదాభిరామ వినుర వేమ

Human Assessment:
The model reproduced the first three verses verbatim and modified only the final verse to fit the input query (Vemana Satakam).

However no such poem exists in Vemana satakam.

Figure 8: Example of GPT-5 output for the Retrieval–Matching Meaning task, showing the model's verbatim reproduction of input verses with minor modification.

Q: ఈ క్రింది పద్యభావానికి సరిపోయే పద్యాన్ని చెప్పండి.
EN: Tell me a poem that fits the following theme.
కుండెడు గాడిదపాలకంటె గరిటెడు ఆవుపాలు మేలు. అలాగే ప్రేమతో పెట్టే భోజనం పెడికెడైనా ఎంతో మంచిది.
EN: A spoonful of cow's milk is better than a pot of donkey's milk. Similarly, a meal served with love, even a small meal, is much better.

GPT-5 Output:
కుండెడు గాడిదపాల కంటె గరిటెడు ఆవుపాలు మేలు,
కొండంత వంటకముల కంటె కొంచెము సత్సారం మేలు;
గుండెనిండ ప్రేమతో వడ్డించిన గింజెడు అన్నమయినా
అందెగదా అమృతసారము, అన్నిటికన్న మేలు.

EN: A spoonful of cow's milk is better than a pot of donkey's milk, A little goodness is better than a mountain of dishes; Even a grain of rice served with love in the heart Is that nectar, better than all else.

Human Assessment:
సత్సారం: There is no such word in Telugu
గింజెడు అన్నమయినా: A grain of rice - The word గింజెడు (grain) is used to refer raw rice but not cooked rice (అన్నము)

అందె: There is no such word in Telugu
అందెగదా అమృతసారము, అన్నిటికన్న మేలు : Even if we take the closest matching words, this information is not provided in the input summary.

Figure 9: Example of GPT-5 output for the Generation–Poem from Summarization task, showing semantic alignment but lexical and contextual inaccuracies in Telugu usage.

Q: పద్యాత్మకమైన పొడుపుకథలను మీ సొంతవాక్యాల్లో చెప్పండి.
EN: Tell poetic riddles in your own words..
కుక్క గురించి ఒక పొడుపుకథ పద్యరూపంలో చెప్పండి.
EN: Tell a riddle about a dog in poem.

GPT-5 Output:
వాసనలే వాకిట దారిచూపే గ్రంథం,
ముక్కుతోనే లోకమంత చదివే పంథం;
ముంగిట కాపాడి దొంగ చప్పుడు ముట్టే,
యజమాని నవ్వితే తోకతాళం కట్టే —
ఈ ఇంటి బంగారు బంట్రోతు నేనెవరు?

EN: Smell is the guidebook at the door,
The path that reads the whole world with just your nose;
The one who guards the front door and stops the thief, The one who locks his tail when the owner laughs — Who am I, the golden guardian of this house?

Human Assessment:
The generated poem lack coherence and contain incorrect sentence structures in Telugu.

యజమాని నవ్వితే తోకతాళం కట్టే (The one who locks his tail when the owner laughs) - This doesn't make any sense.

Figure 10: Example of GPT-5 output for the Generation–Poem (Riddle) task, illustrating incoherent verse construction and incorrect sentence structures in Telugu.

Q: ఈ క్రింది పద్యం చంపకమాల ఛందస్సులో ఉంది. ఈ పద్యభావాన్ని వేరే విధంగా చెప్పండి. క్రొత్త పద్యం కూడా చంపకమాల లక్షణాలతో ఉండాలి.

EN: The following poem is in Champakamala rhyme. Rephrase this poem in a different way. The new poem should also have the characteristics of Champakamala

అనుపమ మెట్టివారలకు నందదు ధర్మపథంబు ధాత్రిలో;
విను, పది వేపురం దొకడు విశ్రుతధర్మపరాయణుండు గ
ల్గునొ కలుగండొ సందియము; గోరి సనాతనధర్మ మూఁది యె
వ్వనికిని నీకుఁ బోలె బుధ వత్సల! ఇట్లు చరింపవచ్చునే

GPT-5 Output:
ఘనముగ ధర్మమార్గమునఁ గాలుని సేవ ఘటించువారెట
ల్గనులని వెదక, నెంతయినఁ గానరు, కోటియుగమ్ము నొంటరే
అనుపమ మైన ధీరతయు నంతియె; నీవయినట్ల శాశ్వత
స్థితిగతి పూజసేయఁగనె జేయగఁజాలరు, మిత్రవత్సలా!

Output from a Chandassu tool
(https://chandamu.github.io/chandam.html)

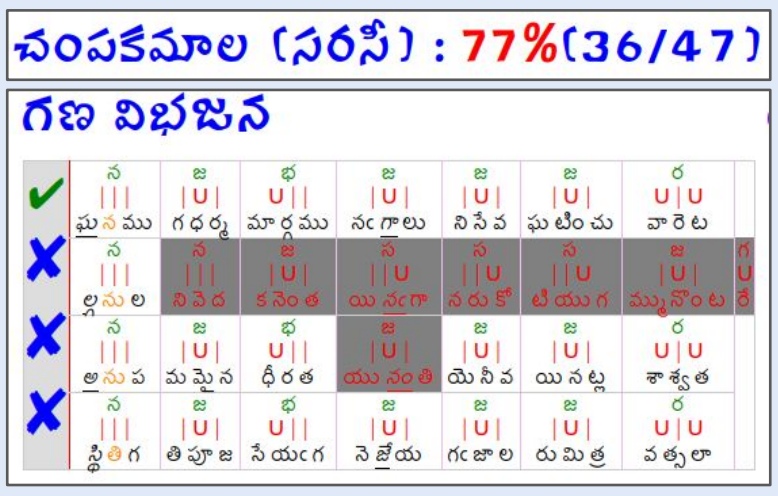

Figure 11: Example of Gemini-2.5-Pro output for the Generation–Style Transfer task, showing partial adherence to the target Champakamala rhyme scheme with approximately 77% metrical accuracy as verified using the Chandam tool.

| Category | Sub-Category | TE Prompt | EN Translation |
|---|---|---|---|
| Retrieval | Morphology | ఈ క్రింది పద్యానికి ప్రతిపదార్థం వ్రాయండి. $INPUT_QUESTION$ | Write an meaning for each word to the following poem. $INPUT_QUESTION$ |
| Retrieval | Matching Meaning | ఈ క్రింది పద్యపాదం పేరు పొందిన శతక పద్యంలోని మొదటి పాదం. ఈ పద్యాన్ని ఉన్నదున్నట్లుగా చెప్పండి. $INPUT_QUESTION$ | The following verse is the first verse of the popular Sataka poem. Recite this poem as it is. $INPUT_QUESTION$ |
| Retrieval | First Verse | ఈ క్రింది పద్యపాదం పేరు పొందిన శతక పద్యంలోని మొదటి పాదం. ఈ పద్యాన్ని ఉన్నదున్నట్లుగా చెప్పండి. $INPUT_QUESTION$ | The following verse is the first verse of the famous sataka poem. Recite this poem as it is. $INPUT_QUESTION$ |
| Retrieval | Middle/Random Verse | ఈ క్రింది పద్యపాదం పేరు పొందిన శతకపద్యంలోని ఒక పాదం. ఈ పద్యాన్ని ఉన్నదున్నట్లుగా చెప్పండి. మీ సొంతవాక్యాలు ఉపయోగించవద్దు. $INPUT_QUESTION$ | The following verse is a verse of the famous Shataka poem. Recite this poem as it is. Do not use your own sentences. $INPUT_QUESTION$ |
| Retrieval | Last Verse | ఈ క్రింది పద్యపాదం పేరు పొందిన శతకపద్యంలోని చివరి పాదం. దీని ముందు పాదాలను గుర్తించి పద్యాన్ని పూరించండి. $INPUT_QUESTION$ | The following verse is the last verse of the famous sataka poem. Identify the verses before it and complete the poem. $INPUT_QUESTION$ |
| Generation | Poem Summary | ఈ క్రింది పద్యం భావమేంటి? $INPUT_QUESTION$ | What is the meaning of the following poem? $INPUT_QUESTION$ |
| Generation | Poem from Summary | ఈ క్రింది పద్యభావానికి సరిపోయే పద్యాన్ని చెప్పండి. $INPUT_QUESTION$ | Give the poem that matches the following verse. $INPUT_QUESTION$ |
| Generation | Riddle Poem from Word | పద్యాత్మకమైన పొడుపుకథలను మీ సొంతవాక్యాల్లో చెప్పండి. $INPUT_QUESTION$ | Recite poetic riddles in your own sentences. $INPUT_QUESTION$ |
| Generation | Poem from a problem | అవధాన ప్రక్రియలో పద్యంలో ఒక పాదాన్ని తార్కిక అంశంతో ముడిపెట్టి వర్ణన చేయమని అడగడాన్ని సమస్య అంటారు. ఈ క్రింద ఇచ్చిన సమస్యను ఆధారంగా చేసుకుని ఒక పద్యం చెప్పండి. $INPUT_QUESTION$ | Recite a poem based on the problem (as in Avadhanam) given below. $INPUT_QUESTION$ |
| Generation | Style transfer | ఈ క్రింది పద్యం $CURRENT_POEM_STYLE$ ఛందస్సులో ఉంది. ఈ పద్యభావాన్ని $NEW_POEM_STYLE$ ఛందస్సులోకి మార్చి చెప్పండి. $INPUT_QUESTION$ | The following poem is in the $CURRENT_POEM_STYLE$ rhyme. Recite this poem in $NEW_POEM_STYLE$ verse. $INPUT_QUESTION$ |
| Support | Error detection and correction | ఈ క్రింది పద్యంలోని తప్పులను (గణము/యతి/ప్రాస/పద్య లక్షణములు) గుర్తించి సవరించండి. $INPUT_QUESTION$ | Identify and correct the errors (meter/syllable/rhyme/verse features) in the following poem. $INPUT_QUESTION$ |
| Support | Vocabulary Suggestion | ఈ క్రింది పద్యపాదంలో ఖాళీయందు $MASKED_WORD$ అనే భావముతో సరిపోయేలా ఉన్న పదాన్ని సూచించండి. పద్యపాదం $INTENDED_POEM_STYLE$ ఛందస్సులో ఉండాలి. $INPUT_QUESTION$ | In the following poem, indicate the word that matches the meaning of $MASKED_WORD$. The poem should be in $INTENDED_POEM_STYLE$ style. $INPUT_QUESTION$ |

Table 5: Overview of prompts used for evaluation, showing Telugu (TE) prompts and their English (EN) translations across retrieval, generation, and support categories